\documentclass[journal]{IEEEtran}

\usepackage{amssymb}
\usepackage{amsmath}
\usepackage{multirow}
\usepackage{graphicx}
\usepackage{booktabs}
\usepackage{color}
\usepackage{bbding}
\usepackage{cite}
\usepackage{url}
\usepackage{algorithm}
\usepackage{algorithmic}

\begin{document}

\title{DaLPSR: Leverage Degradation-Aligned Language Prompt for Real-World Image Super-Resolution}

\author{Aiwen Jiang~\textit{IEEE Member}, Zhi Wei, Long Peng, Feiqiang Liu, Wenbo Li, Mingwen Wang*~\textit{IEEE Member}
\thanks{Aiwen Jiang, Zhi Wei, Feiqiang Liu and Mingwen Wang are with School of Computer and Information Engineering, Jiangxi Normal University, Nanchang, China. Aiwen Jiang and Mingwen Wang are also with School of Digital Industry, Jiangxi Normal University, Shangrao, China. Long Peng is with Institute of Advanced Technology, University of Science and Technology of China, Hefei, China.  Wenbo Li is with Noah’s Ark Lab, Huawei Technologies Ltd., ShenZhen, China.}
\thanks{Manuscript received June, XXX, 2024; revised XXXX,XXXX.}}


\maketitle

\begin{abstract}
Image super-resolution pursuits reconstructing high-fidelity high-resolution counterpart for low-resolution (LR) image. In recent years, diffusion-based models have garnered significant attention due to their capabilities with rich prior knowledge. The success of diffusion models based on general text prompts has validated the effectiveness of textual control in the field of text2image. However, given the severe degradation commonly presented in low-resolution images, coupled with the randomness characteristics of diffusion models, current models struggle to adequately discern semantic and degradation information within severely degraded images. This often leads to obstacles such as semantic loss, visual artifacts, and visual hallucinations, which pose substantial challenges for practical use. To address these challenges, this paper proposes to leverage degradation-aligned language prompt for accurate, fine-grained, and high-fidelity image restoration. Complementary priors including semantic content descriptions and degradation prompts are explored. Specifically, on one hand, image-restoration prompt alignment decoder is proposed to automatically discern the degradation degree of LR images, thereby generating beneficial degradation priors for image restoration. On the other hand, much richly tailored descriptions from pretrained multimodal large language model elicit high-level semantic priors closely aligned with human perception, ensuring fidelity control for image restoration. Comprehensive comparisons with state-of-the-art methods have been done on several popular synthetic and real-world benchmark datasets. The quantitative and qualitative analysis have demonstrated that the proposed method achieves a new state-of-the-art perceptual quality level, especially in real-world cases based on reference-free metrics. Related source codes and pre-trained parameters will be public in \url{https://github.com/puppy210/DaLPSR}.
\end{abstract}


\IEEEpeerreviewmaketitle

\section{Introduction}
\IEEEPARstart{I}{mage} super-resolution pursuits improving image clarity and overall visual quality for low resolution (LR) image\cite{Sun_2024_CVPR,chen2022real,liu2023evaluating,zhang2024seal,peng2024efficient,chen2023human,huang2020unfolding,li2022best,lugmayr2019unsupervised,ji2020real,mou2022metric,liu2022blind,fritsche2019frequency,zhou2020guided,wang2021unsupervised,li2021review}. It involves an intricate task of extracting nuanced perceptual details from LR counterpart to reconstruct high-fidelity and high-resolution image. In recent years, the popularity of deep learning has promoted profound advancements in this field. However, traditional mainstream methods\cite{dong2015image,kim2016accurate,ledig2017photo} often exhibited a propensity towards training on delimited degradation scenarios, thereby constraining their adaptability to unknown complex conditions. As a consequence, real-world applications were often ensnared by the challenges inherent in navigating through diverse and complex degradation scenarios, encompassing denoising, deblurring, and compression artifact removal among other intricacies\cite{zhang2021designing,zhang2023crafting,park2023learning,li2022face,xiao2020degradation,wolf2021deflow,elad1997restoration,peng2024lightweight}.

To recover realistic high-resolution (HR) images with perceptually clear details, some researchers proposed to employ long-short camera lenses to collect LR-HR image pairs from real world\cite{chen2019camera,cai2019toward,wei2020component}. Some other researchers employed more economical approach to simulate complex real-world image degradation using random combinations of basic degradation operations\cite{yang2023synthesizing,bulat2018learn,wu2023datasetdm,sehwag2022generating,maeda2020unpaired}. Representative works in this regard include BSRGAN\cite{zhang2021designing}, Real-ESRGAN\cite{wang2021real}, and their variants\cite{chen2023human,liang2022details,liang2022efficient}.

Recent endeavors have heralded the ascendancy of denoising diffusion probabilistic models\cite{ho2020denoising} within the realm of image generation. Establishing works\cite{rombach2022high,saharia2022photorealistic,wang2024exploiting} have validated that diffusion-based super-resolution models consistently outshone their counterparts relying on generative adversarial networks on various public datasets. More recently, in response to the exigencies posed by the labyrinthine issues of image degradation, particularly in cases where characteristics of degradation remain nebulous, large-scale pre-trained text-to-image models are increasingly inclined to serve as beneficial helpers\cite{mou2024t2i}. In this routine, ControlNet\cite{zhang2023adding} introduced as an innovative adaptor can effectively utilize supplementary conditions to steer the generative capabilities of pretrained models. Prior research such as DiffBIR\cite{lin2023diffbir} has demonstrated that integrating ControlNet into image super-resolution process yields remarkable fidelity in reproducing realistic details.

Nevertheless, it poses challenges to derive semantic priors in the form of textual prompts into pretrained text-to-image models, given the complex degradation patterns observed in LR images. PASD\cite{yang2023pixel}  and SeeSR\cite{Wu_2024_CVPR} proposed to employ existing labeling models to extract object labels as high-level prompts. Unfortunately, these annotation cues often lack intricate details for scene comprehension. Moreover, substantial degradation in LR image, such as disruption of local structures, may lead to semantic ambiguities, causing the reconstructed HR image to potentially exhibit semantic inaccuracies, thereby deteriorating final super-resolution performance.

Multimodal large language models (MLLMs) \cite{zhang2023llama,li2023blip,liu2024visual,zhu2023minigpt} have showcased remarkable proficiency in visual comprehension, and obtained remarkable success across a gamut of downstream tasks. SUPIR\cite{Yu_2024_CVPR} was one of the few latest pioneering works that empowered the capabilities of MLLMs for super-resolution task. It initially integrated robust features from LR images into LDM image decoder to produce a HR reference image. Subsequently, it leveraged MLLMs to generate precise and detailed textual descriptions based on the HR reference. Finally, the generated descriptions acted as semantic prompt to guide pretrained SDXL\cite{podell2023sdxl} for image restoration. As it claimed that it was the largest-ever image restoration method, inevitably, it substantially escalated computational resource demands, which were unaffordable for ordinary applications.

We deem that accurate and comprehensive prompts are crucial in T2I based super-resolution schemes. Therefore, in this paper, we have proposed an effective multimodal framework (DaLPSR) that leverages degradation-aligned language prompt for real-world image super-resolution task. In the proposed DaLPSR, two complementary priors are generated to overcome preceding mentioned problems. Specifically, on one hand, image-restoration prompt alignment decoder (IRPAD) is proposed and trained to automatically discern the degradation degree of LR images. To facilitate the generation of image-restoration prompt, we have constructed an auxiliary dataset with triplet data that signifies the degradation process. We adopt to discretize degradation degree into several intervals, and take the generation process of degradation prompt as fine-grained retrieval process. The generated prompts from IRPAD provide degradation priors for Stable Diffusion. On the other hand, MLLM is utilized to acquire high-level semantic priors, ensuring fidelity control for the generated content. To ensure that the high-level semantic priors generated by MLLM correlate closely with semantic ground-truth, along with fine-tuning image encoder for feature coherence between LR and HR image, we introduce Recognize Anything Model (RAM)\cite{zhang2024recognize} to generate object labels as part of prompt instruction for MLLM, guiding it to generate accurate high-level semantic priors aligned with image subjects.

In summary, the main contributions of this work are as followings:
\begin{enumerate}
\item We have proposed an effective and computational affordable one-stage degradation-aligned multimodal framework for real-world image super-resolution task. Complementary priors including semantic content descriptions and degradation prompts are leveraged for accurate, fine-grained, and high-fidelity image restoration. Much richly tailored prompts for MLLMs elicit high-level semantic priors closely aligned with human perception.
\item We have proposed an effective image-restoration prompt alignment decoder that is capable of automatically identifying degradation level of LR images, thereby generating beneficial degradation priors for image restoration. To the best of my knowledge, we are the first to explicitly concern degradation prompts for image super-resolution.
\item We have engineered an image-restoration prompt generation pipeline that seamlessly incorporates textual degradation prompts into SR datasets. As a result, we have innovatively contributed an auxiliary triplet dataset for degradation learning.
\item We have conducted comprehensive experiments on popular benchmark datasets, including both synthetic and real-world cases. The experiment results according to reference-based and reference-free metrics have demonstrated that the proposed method achieves a new state-of-the-art perceptual performance on real-world cases.
\end{enumerate}

\section{Related Work}
\subsection{GAN-based Real-ISR methods}
Ever since the emergence of SRCNN\cite{dong2014learning}, deep learning based image super-resolution (ISR) has garnered significant attention in computer vision area. Various methods aiming at enhancing the reconstruction precision of ISR have been proposed. Among these methods, most of them generally operated on predetermined degradation that typically employed bicubic downsampling to generate low-resolution image. As a result, when they confronting intricate and unpredictable degradation in real-world scenarios, such straightforward degradation assumption would constrain their efficacy.

Recent studies emphasized the imperative of delving into more intricate degradation models to better emulate real-world conditions. For instance, BSRGAN\cite{zhang2021designing} introduced a stochastic degradation modeling strategy. Real-ESRGAN\cite{wang2021real} adopted sophisticated high-order degradation strategy. By leveraging training datasets featuring more realistic degradation, both BSRGAN and RealESRGAN harness generative adversarial networks (GANs) to reconstruct requisite high-resolution images.

Albeit yielding more authentic details, the instability inherent in training GANs often leads to ISR outputs plagued by unnatural visual artifacts. Being aware the fact, subsequent endeavors such as LDL\cite{liang2022details} and DeSRA\cite{xie2023desra} concentrated on mitigating these artifacts. Specifically, LDL proposed to explicitly discriminate visual artifacts from realistic details, and design locally discriminative learning framework to regularize the adversarial training. DeSRA proposed to locate problematic areas based on a defined relative local variance distance and semantic-aware thresholds. After detecting the artifact regions, they proposed to develop a finetune procedure to improve GAN-based SR models. Nevertheless, they still frequently encountered challenges in generating more naturally plausible details.

\subsection{Diffusion-based Real-ISR methods}
The diffusion model is a type of generative model that utilizes diffusion ideas from physical thermodynamics, mainly including forward diffusion and backward diffusion processes. In the realm of image generation, denoising diffusion probabilistic model (DDPM)\cite{ho2020denoising} stood as a pioneering effort to employ diffusion models, boasting commendable performance and surpassing GAN-based methods. As following work after DDPM, denoising diffusion implicit model (DDIM)\cite{song2020denoising} introduced a distinct yet effective sampling technique to address the inference inefficiency of traditional diffusion process. Latent diffusion model (LDM)\cite{rombach2022high} from another perspective extended diffusion models into latent space of pre-trained autoencoders, enabling training large-scale model with limited computational resources. Concurrently, in the domain of text-to-image field, Stable Diffusion (SD), based on latent diffusion, have garnered significant success.

In recent times, diffusion models have been astutely repurposed in the field of Real-ISR to be leveraged as priors. There are two main categories for diffusion-based SR methods. The first category\cite{xia2023diffir,yue2024resshift,wang2023sinsr,peng2024towards} involves training diffusion model from scratch, where models use a combination of LR images and noise as input at each diffusion step. The second category\cite{wang2024exploiting,lin2023diffbir,yang2023pixel,Wu_2024_CVPR} leverages prior knowledge from pre-trained diffusion models (e.g., Stable Diffusion). Methods of the second category required implementing "relearning" of large models through adaptors such as ControlNet\cite{zhang2023adding}, LoRA\cite{hu2022lora}, etc. Specifically, StableSR\cite{wang2024exploiting} fine-tuned Stable Diffusion (SD) model and employed feature warping to strike balance between fidelity and perceptual quality. DiffBIR\cite{lin2023diffbir} adopted a two-stage methodology, which reconstructed an initial HR estimate followed by utilizing SD to refine and augment visual details. Objectively speaking, under the impetus of diffusion-based methods, significant progress has been made on image super-resolution, especially on precise restoration of image details. However, it still presents great challenge when facing restoration of LR images featuring complex distortions and spatially intertwined object locations.

\subsection{Semantic Guidance-based Real-ISR methods}
Benefiting from robust generative priors, pretrained text-to-image diffusion models are becoming increasingly popular for addressing Real-ISR challenges. Previous ISR works have attempted to leverage existing visual perception models (e.g., ResNet\cite{he2016deep}, BLIP\cite{li2022blip}, or RAM\cite{zhang2024recognize}) to comprehend image content and provide multimodal semantic priors as prompts. Typically, PASD\cite{yang2023pixel} and SeeSR\cite{Wu_2024_CVPR} introduced high-level semantic information (e.g., object labels) as additional conditions, to harness the potential of pre-trained text-to-image diffusion models. Nevertheless, these prompts were often insufficient since they merely contained basic information from object recognition, lacking details of spatial localization and scene understanding.


\begin{figure*}[!t]
	\centering
	\includegraphics[width= 1\textwidth]{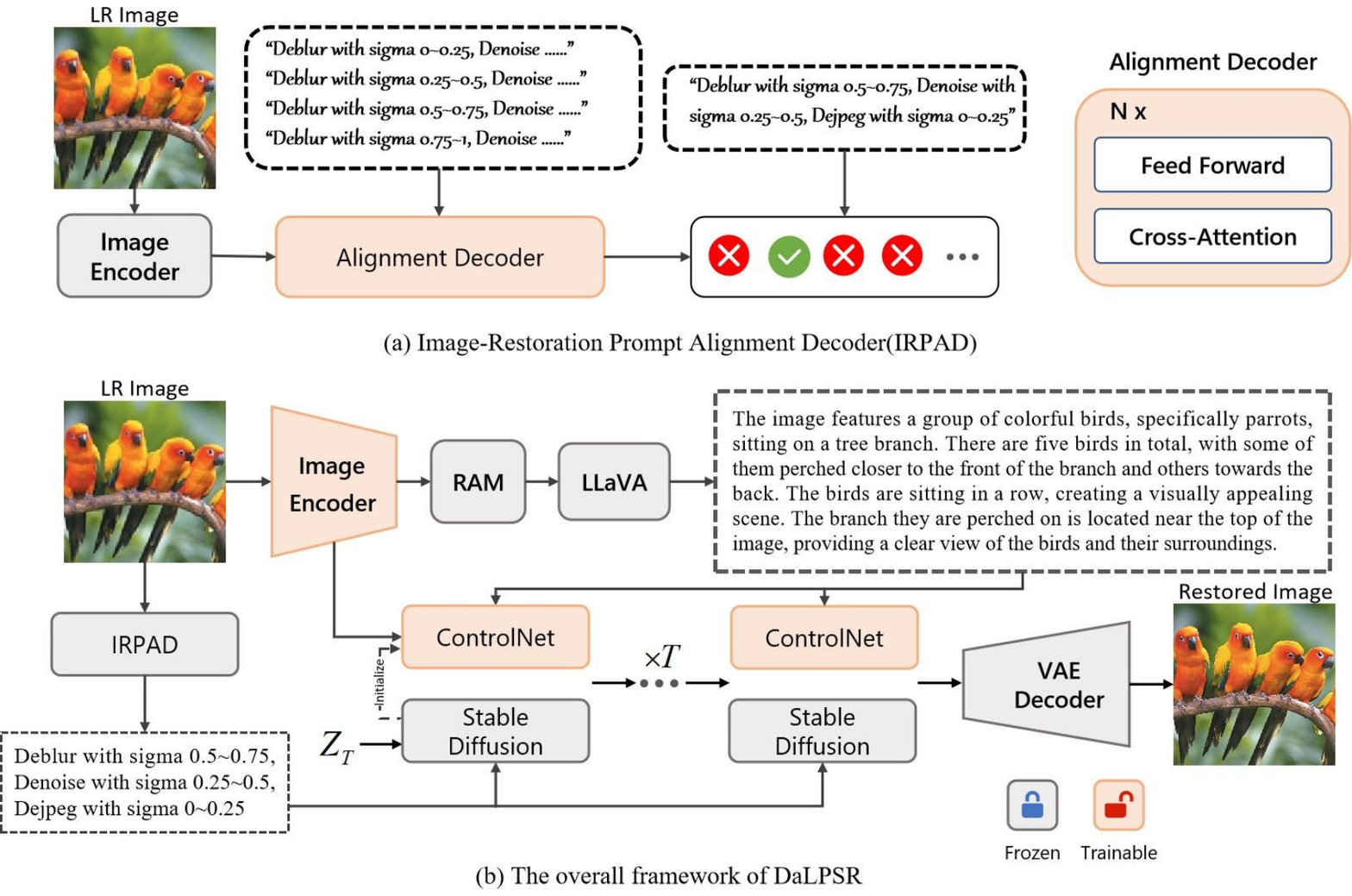}
	\caption{Overview of the proposed method. (a) Image-Restoration Prompt Alignment Decoder (IRPAD) is designed to extract image-restoration prompts from LR images. (b) The extracted prompts from IRPAD are integrated into image generation process, enabling the model to leverage semantic priors and restoration priors to enhance the quality and fidelity of the reconstructed HR images.}
	\label{fig:overview}
\end{figure*}

Benefiting from scale expanding of multimodal training data and success of large language model (LLMs), multimodal LLM (MLLMs) have demonstrated significant capabilities in semantic understanding. They are empowered to well capture and interpret complex semantic relationships across different modalities, leading to more accurate and contextually relevant outputs. However, in the current usage of MLLMs, it is still not an easy task to determine appropriate instructions to guide MLLMs to generate outputs aligning with human preference, since crafting these instructions involves not only understanding the nuances of human language and preferences but also how to translate these nuances into machine-understandable prompts. This process necessitates sustained practice and in-depth research to optimize the design of instructions, enabling them to effectively direct large model to generate high-quality, semantically consistent outputs.

\section{Proposed Method}
In text-to-image research field, popular diffusion model primarily leverages semantic textual prompt to generate images that are both more expressive and semantically meaningful. In this study, we deem that it is viable way to enhance model's capability on image super-resolution through utilizing textual prompt as semantic prior during image generation process. Through the incorporation of textual prompt, models can achieve a more precise understanding of semantic content presented in LR image, leading to reconstruction of more clearer super-resolution image with enriching perceptual details.

In this section, we primary describe our proposed DaLPSR network in details. The overview of the proposed network is provided in Figure~\ref{fig:overview}.

\subsection{Preliminary}
\subsubsection{Stable Diffusion}
Stable Diffusion is a latent diffusion model comprising an auto-encoder with Encoder $\mathcal{E}$ and Decoder $\mathcal{D}$, a prompt-conditioning denoising U-Net $\mathcal{E}_{\theta}$ and a CLIP text-encoder $\mathcal{T}_{E}$. Input image $x$ undergoes perceptual compression by Encoder $\mathcal{E}$ into a lower-dimensional latent space $z$. Decoder $\mathcal{D}$ is responsible for converting the latent representation back to pixel space. Text-encoder $\mathcal{T}_{E}$ generates latent text embeddings $\tau$ for input text prompts. Denoising Network $\mathcal{E}_{\theta}$ is employed to gradually process/diffuse information transformed into latent space.

The training of denoising U-Net $\mathcal{E}_{\theta}$ involves minimizing a defined loss function as shown in Equation~\ref{eq:ldm}:
\begin{equation}
\label{eq:ldm}
\mathrm{L}=\mathbb{E}_{z,\tau,\epsilon\sim\mathcal{N}(0,I),t}[ ||\epsilon-\mathcal{E}_\theta(z_t,t,\tau)||_2^2 ],
\end{equation}
where $t$ is diffusion timestep, $z_t$ represents a fully noised image or encoded image generated through a gradual addition of Gaussian Noise $\epsilon\sim\mathcal{N}(0,I)$, and $\tau$ is latent text embedding.

Stable diffusion employs cross-attention to incorporate textual prompts into image generation process. This process involves encoding latent image features $\mathit{F_{img}}$ and textual features $\mathit{F_\tau }$. It maps respective features to query $Q={{W}_{Q}}\cdot {{F}_{img}}$, key $K={{W}_{K}}\cdot {\mathit{F}_{\tau }}$, and value $V={{W}_{V}}\cdot {\mathit{F}_{\tau }}$ through linear projection layers, where $W_Q$, $W_K$, $W_V$ respectively represent the weight parameters of the query projection layer, key projection layer and value projection layer. The calculation of cross attention is subsequently executed by performing a weighted summation of the value features $V$. The formula of cross-attention is as shown in Equation~\ref{eq:ca}.
\begin{equation}
\label{eq:ca}
Attention(Q,K,V)=Softmax(\frac{Q{{K}^{T}}}{\sqrt{d}})V,
\end{equation}
where $d$ represents the output dimension of key and query features. The output of cross-attention block is utilized to update the latent features in Stable-Diffusion model.

\subsubsection{ControlNet}
\begin{figure}[h]
	\centering
	\includegraphics[width=0.4\textwidth]{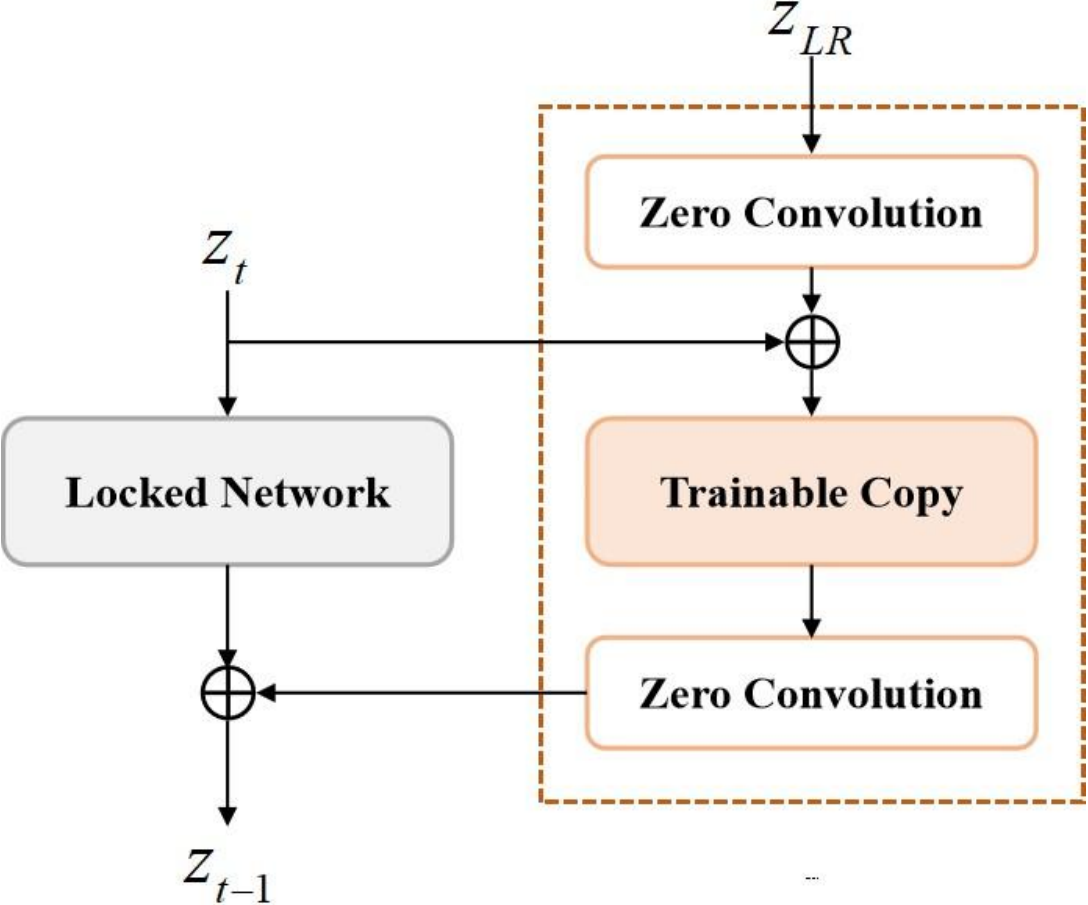}
	\caption{The fundamental structure of ControlNet. $Z_{\text{LR}}$ is the latent representation of LR image.}
    \label{fig:controlnet}
\end{figure}

ControlNet\cite{zhang2023adding} is specially developed for efficiently managing pretrained large diffusion models while accommodating additional input conditioning. It involves creating two distinct copies of the weights from pretrained diffusion model. One is a trainable copy and the other is a locked copy. The trainable copy is trained on task-specific datasets to learn some required condition, while the locked copy remains fixed. These two copies are connected through zero-convolutions, enabling efficient training that is in comparable speed with fine-tuning, while avoiding introducing additional noise to deep features of original pretrained network. The fundamental architecture of ControlNet is shown in Figure~\ref{fig:controlnet}.

\subsection{Image-Restoration Prompt Alignment Decoder (IRPAD)}
\subsubsection{Image-Restoration Prompt Generation Pipeline}
To facilitate the generation of image-restoration prompt, we construct an auxiliary dataset that signifies the degradation process. In the constructed dataset, each data is a triplet [$I_{\text{HR}}$, $I_{\text{LR}}$, $p_r$], where $p_r$ signifies the textual prompt characterizing degradation process, while $I_{\text{HR}}$ and $I_{\text{LR}}$ respectively represents HR and LR image in pair.

\begin{figure}[h]
	\centering
	\includegraphics[width=0.5\textwidth]{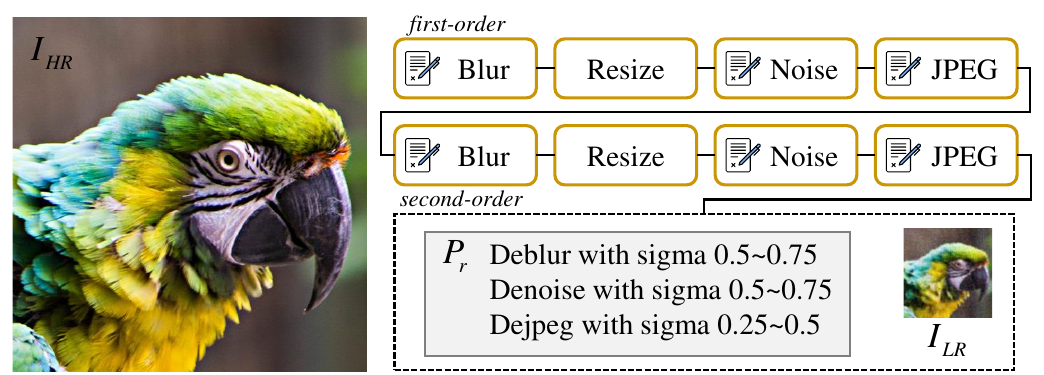}
	\caption{The image-restoration prompt generation pipeline. Discretization strategy is employed on degradation level representation to delineate high-order degradation process. }
	\label{fig:irpad}
\end{figure}

Specifically, the generation pipeline of image-restoration prompt is outlined as in Figure~\ref{fig:irpad}. It consists of two primary components. One is the degradation module and the other is the restoration prompt representation module. The former is responsible for producing HR-LR image pairs. It follow the second-order degradation process as in Real-ESRGAN\cite{wang2021real} to model practical degradations. The latter delineates the degradation process to generate corresponding textual prompts.

Since degradation space is continuous, it is difficult to accurately identify the true degree value for a specific degradation. For simplicity and practical use, in this paper, we adopt a strategy that involves discretizing degradation space into several intervals. As a result, textual descriptions are facilitated to comprehensively and informatively capture nuanced variations and intricacies in degradation process. As shown in Figure~\ref{fig:irpad}, we specifically take 4 intervals, e.g. [0,0.25], (0.25,0.5], (0.5,0.75], and (0.75,1] for degradation degree. Consequently, restoration prompts are generated as descriptions "\textit{Deblur with sigma $0.5\sim0.75$, Denoise with sigma $0.25\sim0.5$, Dejpeg with sigma $0\sim0.25$}". We deem that it is essential for ill-reverse applications like image restoration, helping narrow semantic gap. The adopted strategy can ensure clarity and precision in conveying the intricacies of degradation operations in textual way.

\begin{figure*}[!t]
	\centering
	\includegraphics[width= 1\textwidth]{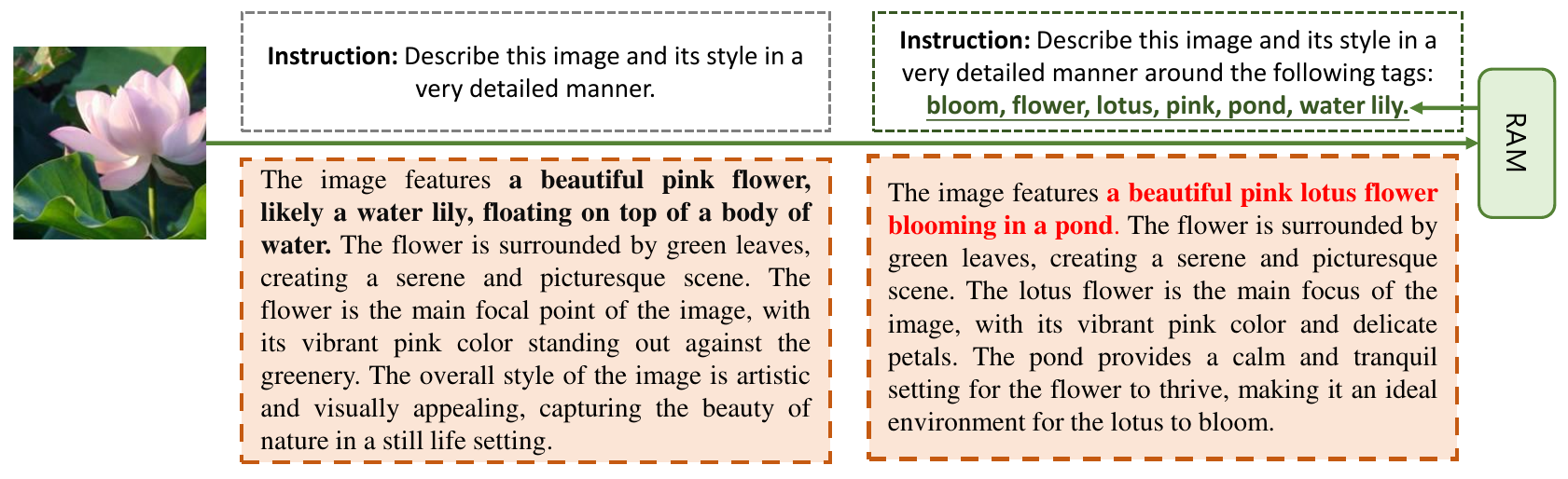}
	\caption{The example demonstrates LLaVA has the capacity to produce high-level semantic prompts that align with human perceptual comprehension through application of tailored prompt instructions. RAM is employed to detect semantic tags to formulate tailored instructions for LR images. }
	\label{fig:MLLM}
\end{figure*}

\subsubsection{Image-Restoration Prompt Alignment Decoder}
Figure~\ref{fig:overview}(a) delineates the image-prompt alignment process. The alignment decoder comprises two successive blocks, each of which is consist of a cross-attention layer and a feedforward layer. Embeddings of degradation prompts are aligned with image features within the alignment decoder.

The alignment decoder takes degradation prompts and image features as inputs.  In the cross-attention layer, these degradation prompts act as queries, while the image features serve as keys and values. During the pretraining stage, the process results in an output that is the degradation prompt most closely aligned with the input image, ensuring precise relevance and alignment.

Throughout the training process of the alignment decoder, the task is structured as a fine-grained retrieval problem. In the pretraining phase, the model is responsible for selecting the most pertinent degradation prompt for each degradation type from the predefined intervals.  This involves learning to effectively align degradation prompts with image features, enabling the retrieval of the most relevant prompts based on the provided image features.  Consequently, this approach allows us to optimize computational efficiency, highlight the significance of degradation prompts, and maintain scalability for accommodating additional types of degradation in the future.

\subsection{Multi-Modality Semantic Priors Guidance}
By significantly scaling up the size and complexity of both data and language models, multimodal large language models (MLLMs) have achieved remarkable performance in high-level tasks like image captioning. In our work, we utilize a cutting-edge MLLM called LLaVA \cite{liu2024visual} to enhance the perception of low-resolution (LR) images and extract valuable semantic information. LLaVA's advanced capabilities allow it to capture complex relationships and patterns within visual data, making it exceptionally effective for this purpose.

To improve the alignment between the semantic content of LR images and the descriptions generated by LLaVA, we propose enhancing the feature extraction capability of the image encoder $\mathcal{E}$ for LR images. This is achieved by fine-tuning the image encoder $\mathcal{E}$ to minimize the loss function $\mathcal{L}{\mathcal{E}} = |\mathcal{E}(I{\text{LR}})-\mathcal{E}(I_{\text{HR}})|_{2}^{2}$, thereby reducing the differences between LR and high-resolution (HR) images. This process ensures that the encoder extracts features from LR images that are closer to those of HR images, leading to more accurate semantic interpretations.

Furthermore, providing clear and comprehensive instructions to MLLMs is crucial to ensuring that their outputs align with human intentions and preferences. Given the powerful capabilities of RAM \cite{zhang2024recognize} in image tag recognition, we employ RAM to detect relevant tags for LR images. These identified tags are then used to create specific instructions tailored to each LR image. As illustrated in Figure~\ref{fig:MLLM}, the tags extracted from an LR image become part of the instructive prompts for LLaVA. By utilizing these targeted prompt instructions, LLaVA can produce high-level semantic image descriptions that resonate with human perceptual understanding, thereby enhancing the coherence and relevance of the generated content.

\subsection{Training Strategy}

The training strategy of the proposed method is delineated in Algorithm 1. During model training, we initially derive the latent representation $\mathbf{z}_0$ of HR image. Subsequently, we incrementally introduce noise to this latent representation, producing a noisy latent $\mathbf{z}_t$, where $t$ represents a randomly sampled diffusion step. Considering various conditions such as the diffusion step $t$, latent LR input $\mathbf{z}_{\text{LR}}$, restoration prompt $p_r$, and high-level semantic prompt $p_s$, we train the network $\epsilon_{\theta} $ to predict the noise introduced into the noisy latent $\mathbf{z}_t$. The optimization objective is:
\begin{equation}
\label{eq:op}
\mathcal{L}=\mathbb{E}_{z_0,z_{\text{LR}},t,p_r,p_s,\epsilon\sim\mathcal{N}}[\|\epsilon-\epsilon_\theta (\mathbf{z}_t, \mathbf{z}_{\text{LR}},t,p_r,p_s) \|_2^2].
\end{equation}

\begin{algorithm}
\small
\caption{Training}
\label{alg:Traning}
\begin{algorithmic}
\STATE \textbf{Training datasets: }$(I_{\text{HR}}, I_{\text{LR}}, p_r, p_s\sim{MLLM(I_{\text{LR}}))}$
\STATE \textbf{sample: }$\mathbf{z}_{0},\mathbf{z}_{\text{LR}}\sim(I_{\text{HR}}, I_{\text{LR}})$
\WHILE{not converged}
    \STATE $t\sim\text{Uniform}(\{1,\ldots,T\})$
    \STATE
    \textbf{sample: }$\epsilon\sim\mathcal{N}(\mathbf{0},\kappa^2\eta_t\mathbf{I})$
    \STATE $\mathbf{z}_t=\mathbf{z}_{0}+\epsilon $
    \STATE The optimization objective is:
    \STATE $\mathcal{L}=\mathbb{E}_{z_0,z_{\text{LR}},t,p_r,p_s,\epsilon\sim\mathcal{N}}[\|\epsilon-\epsilon_\theta (\mathbf{z}_t,\mathbf{z}_{\text{LR}},t,p_r,p_s) \|_2^2]$
\ENDWHILE
\end{algorithmic}
\end{algorithm}

\section{Experiments}
\subsection{Datasets and metrics}
\subsubsection{Training datasets}
As similar settings to conventional training strategy, our training data are collected from DIV2K\cite{agustsson2017ntire}, DIV8K\cite{gu2019div8k}, Flickr2K\cite{timofte2017ntire}, OST\cite{wang2018recovering}, and a subset of top 10K images from FFHQ\cite{karras2019style}. Initially, HR images are randomly cropped to a size of $512\times512$. Then, image-restoration prompt generation pipeline is employed to produce training dataset containing HR-LR image pairs and corresponding restoration prompts. Subsequently, the proposed network is trained using the generated training data.

\subsubsection{Test Datasets} 
In order to thoroughly and reliably evaluate the proposed model, we conducted experiments on several synthetic and real-world benchmark datasets that widely adopted in super-resolution tasks.

In case of synthetic dataset, we utilized the validation set from DIV2K\cite{agustsson2017ntire} dataset. Following the same experiment settings as SeeSR\cite{Wu_2024_CVPR}, we utilize the randomly cropped 3K image patches for testing\footnote{https://huggingface.co/datasets/Iceclear/StableSR-TestSets}. The HR images are of size $512\times512$. The corresponding LR images of size $128\times128$ are generated using the same degradation pipeline as training stage.

\begin{table*}[!t]
\footnotesize
\caption{Quantitative evaluation against state-of-the-art methods on real-world benchmarks. The best performances across each metric are highlighted in red and bold.}
\label{tab:comparisonSOTA}
\tabcolsep 0.8pt 
\begin{tabular*}{\textwidth}{ccccccccccccc}
\toprule
  Datasets & Metrics & BSRGAN & Real-ESRGAN & LDL & FeMaSR & StableSR & ResShift & PASD & DiffBIR & SeeSR & SUPIR* & Ours \\
 &  & ICCV 2021 & ICCV 2021 & CVPR 2022 & MM 2022& IJCV2024 & NeurIPS 2023& ECCV 2024& arxiv 2023& CVPR 2024 & CVPR 2024 \\\hline
  & LPIPS $\downarrow $  & 0.4136  & \textcolor{red}{\textbf{0.3868}} & 0.3995 & 0.3973 & 0.4055   & 0.4284   & 0.4410 &  0.4270 & 0.3876 & 0.3933 & 0.4085 \\
 & FID   $\downarrow $  & 64.28 & 53.46  & 58.94 & 53.70 & 36.57 & 55.77 & 40.77 &  40.42 & 32.79 & \textcolor{red}{\textbf{31.73}} & 34.92  \\
 DIV2K-Val & MANIQA  $\uparrow $ & 0.4834 & 0.5251  & 0.5127 & 0.4869 & 0.5914  & 0.5232  & 0.6049 & 0.6205 & \textcolor{red}{\textbf{0.6236}} & 0.5080 & 0.6146  \\
 & MUSIQ  $\uparrow $ & 59.11  & 58.64  & 57.90  & 58.10  & 62.95  & 58.23 & 66.85  &  65.23 &
 68.29 &  59.68 &\textcolor{red}{\textbf{70.70}} \\
 & CLIPIQA $\uparrow $ & 0.5183 & 0.5424 & 0.5313 & 0.5597 & 0.6486 & 0.5948   & 0.6799 &  0.6664 & 0.6834 & 0.6737 & \textcolor{red}{\textbf{0.7556}}  \\
\hline
 & LPIPS $\downarrow $  & \textcolor{red}{\textbf{0.2670}}  & 0.2727 & 0.2766 & 0.2942 & 0.3018   & 0.3460   & 0.3435 & 0.3658  & 0.3133 & 0.3788 & 0.3952  \\
 & FID   $\downarrow $  & 141.28 & 135.18  & 142.71 & 141.05 & \textcolor{red}{\textbf{128.51}}   & 141.71   & 129.76 & 128.99 & 134.90 & 133.17 & 148.65  \\
 RealSR & MANIQA  $\uparrow $ & 0.5399 & 0.5487  & 0.5485 & 0.4865 & 0.6221   & 0.5285 & 0.6493
 & 0.6253  & \textcolor{red}{\textbf{0.6506}} & 0.5673 & 0.6303  \\
 & MUSIQ  $\uparrow $ & 63.21  & 60.18  & 60.82  & 58.95  & 65.78    & 58.43    & 68.69  & 64.85   & 69.67 & 65.41 & \textcolor{red}{\textbf{71.48}}   \\
 & CLIPIQA $\uparrow $ & 0.5001 & 0.4449 & 0.4477 & 0.5270 & 0.6178   & 0.5444   & 0.6590 & 0.6386  & 0.6630 & 0.6896 & \textcolor{red}{\textbf{0.7199}}  \\
\hline
& LPIPS $\downarrow $  & 0.2883 & 0.2847 & \textcolor{red}{\textbf{0.2815}} & 0.3169 & 0.3315   & 0.4006   & 0.3931 & 0.4599  & 0.3299 & 0.4180  & 0.4048  \\
& FID   $\downarrow $  & 155.63 & \textcolor{red}{\textbf{147.62}}   & 155.53 & 157.78 & 148.98 & 172.26   & 159.24 & 166.79  & 151.88 & 164.64 & 164.66  \\
DRealSR & MANIQA $\uparrow $  & 0.4878 & 0.4907 & 0.4914 & 0.4420 & 0.5591   & 0.4586   & 0.5850 & 0.5923 & \textcolor{red}{\textbf{0.6005}} & 0.4776 & 0.5733  \\
& MUSIQ $\uparrow $  & 57.14  & 54.18 & 53.85  & 53.74  & 58.42    & 50.60 & 64.81  & 61.19   & 65.11 & 58.70 & \textcolor{red}{\textbf{65.88}}  \\
& CLIPIQA $\uparrow $ & 0.4915 & 0.4422 & 0.4310 & 0.5464 & 0.6206   & 0.5342 & 0.6773  & 0.6346  & 0.6708 & 0.6701 & \textcolor{red}{\textbf{0.7245}} \\
\hline
& MANIQA $\uparrow $ & 0.5462 & 0.5582 & 0.5519 & 0.5295 & 0.5841 &0.5417 &0.6066 &0.5902 & 0.6198 & 0.4795 & \textcolor{red}{\textbf{0.6339}}\\
RealLR200& MUSIQ $\uparrow $  & 64.87 & 62.94 & 63.11 & 64.14 & 63.30 & 60.18 & 68.20 & 62.06 & 69.52 & 66.44 & \textcolor{red}{\textbf{72.78}} \\
& CLIPIQA $\uparrow $ & 0.5679 & 0.5389 & 0.5326 & 0.6522 & 0.6068 &0.6486 & 0.6797 & 0.6509 & 0.6814 & 0.6275 & \textcolor{red}{\textbf{0.7728}}\\
\bottomrule
\end{tabular*}
\end{table*}

In case of real-world dataset, we employ three benchmark datasets, DRealSR\cite{wei2020component}, RealSR\cite{cai2019toward} and RealLR200\cite{Wu_2024_CVPR} for evaluation. Except for RealLR200, all datasets undergo cropping HR image to dimensions of $512\times512$ and subsequent LR images of size $128\times128$. LR images in RealLR200 dataset are of various sizes for practical use.

\subsubsection{Compared Methods}
We perform benchmark comparisons for our proposed method against several representative cutting-edge Real-ISR methods. The compared methods are categorized into two primary groups. One group contains GAN-based methods, such as BSRGAN\cite{zhang2021designing}, Real-ESRGAN\cite{wang2021real}, LDL\cite{liang2022details}, FeMaSR\cite{chen2022real}. The other group encompasses diffusion-based methods, like StableSR\cite{wang2024exploiting}, ResShift\cite{yue2024resshift}, PASD\cite{yang2023pixel}, DiffBIR\cite{lin2023diffbir}, SeeSR\cite{Wu_2024_CVPR} and SUPIR\cite{Yu_2024_CVPR}.


\subsubsection{Evaluation indicators}
Both established reference-based and reference-free metrics are utilized for evaluation.

The reference-based perceptual quality assessment is conducted using LPIPS\cite{zhang2018unreasonable} and FID\cite{heusel2017gans} metrics. LPIPS gauges perceptual similarity between the restored images and their references. FID quantifies the distributional dissimilarity between the restored images and their references, providing insights into the fidelity of image restoration. Notably, traditional pixel-level evaluation metrics such as PSNR and SSIM\cite{wang2004image} are excluded from our assessment due to their limited efficacy in capturing nuanced aspects of human perceptual quality.

For reference-free evaluation, we leverage MANIQA\cite{yang2022maniqa}, MUSIQ\cite{ke2021musiq}, and CLIPIQA\cite{wang2023exploring} metrics. These metrics offer a holistic assessment of image quality without relying on reference images, thus providing a more robust evaluation framework for assessing the fidelity and perceptual quality of generated images.

\subsection{Implementation details}
The SD 2.1-base model\footnote{https://github.com/Stability-AI/stablediffusion} servers as the foundational text-to-image model in our experiment. Adam with batch size of 4 is employed as optimizer for training. We configure the learning rate to 2e-5 and fine-tune complete model over 100,000 iterations. During inference, we incorporate LR images with initial noise to establish an enhanced starting point for diffusion process. All training experiments are conducted on image patch of $512\times512$ resolution on NVIDIA 3090 GPU.

\subsection{Comparison with state-of-the-art methods}
\subsubsection{Quantitative Comparisons}
The quantitative comparisons across both synthetic and real-world datasets are presented in Table~\ref{tab:comparisonSOTA}. From the experiment results, we can observe that the proposed method consistently outperforms competing methods across multiple datasets, demonstrating superior performance in terms of various perceptual metrics. Specifically, on RealLR200 dataset, our MANIQA score exhibits a superiority of 2.27 \% over the runner-up method, while our MUSIQ score surpasses the second-best method by 4.69\%, and our CLIPIQA score excels by 13.41\% compared to the next best method.

Notably, compared with the most closely work such as SUPIR, the proposed DaLPSR has several crucial differences. (1) Overall, DaLPSR is one-stage framework while SUPIR utilized two-stage strategy. (2) The priors that SUPIR utilized are only the generated descriptions by MLLM based on initially estimated HR reference. In contrast, DaLPSR, on one hand, with assistance of RAM, employs MLLM directly on encoded LR images to generate semantic priors. As evidenced in the experiments, the generated semantic priors in DaLPSR are more accurate and comprehensive. On the other hand, DaLPSR accounts for degradation alignment, and learns degradation information as complementary degradation priors. (3) The computation burdens of DaLPSR are much less than SUPIR, which enables DaLPSR be more applicable in practical scenario.

We have also noted that GAN-based methods generally exhibit better performance over diffusion-based approaches in reference-based metrics. We speculate that diffusion model possesses capability to generate more intricate texture details, albeit at the expense of marginal reduction in fidelity. Nevertheless, compared to SOTA diffusion-based methods, the proposed method achieves more superior scores in reference-free metrics. Especially on RealLR200 dataset, it attains highest scores across all reference-free metrics, thereby establishing itself as a valuable asset for practical deployment.

\begin{figure*}[!t]
	\centering
	\includegraphics[width= 1\textwidth]{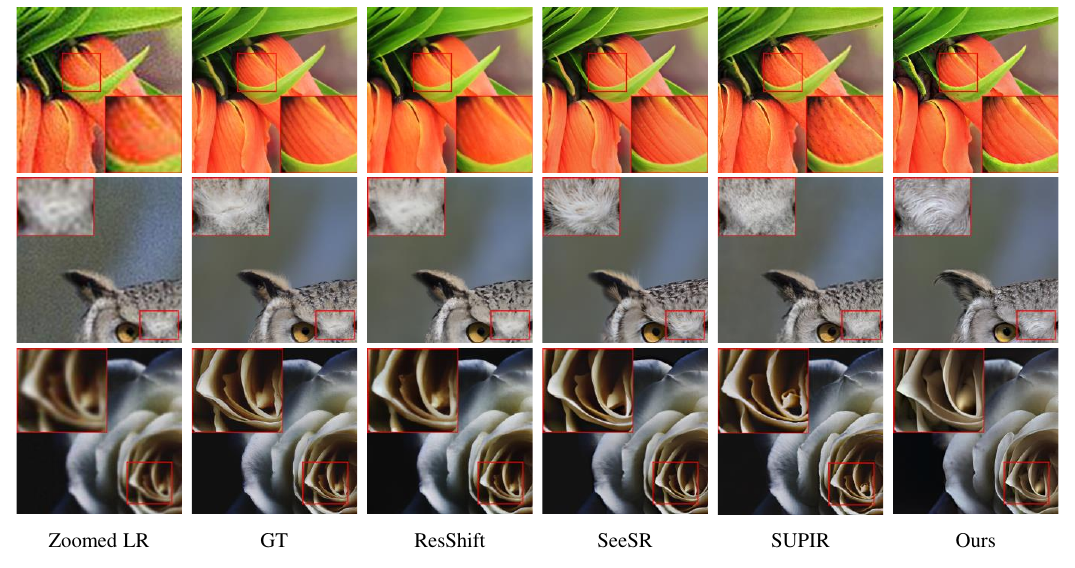}
	\caption{Visual comparisons between the proposed model and other state-of-the-art methods on DIV2K-Val dataset. For a clearer and more detailed view, please zoom in on the images.}
	\label{fig:visualcomp-DIV2k}
\end{figure*}

\begin{figure*}[!t]
	\centering
	\includegraphics[width= 1\textwidth]{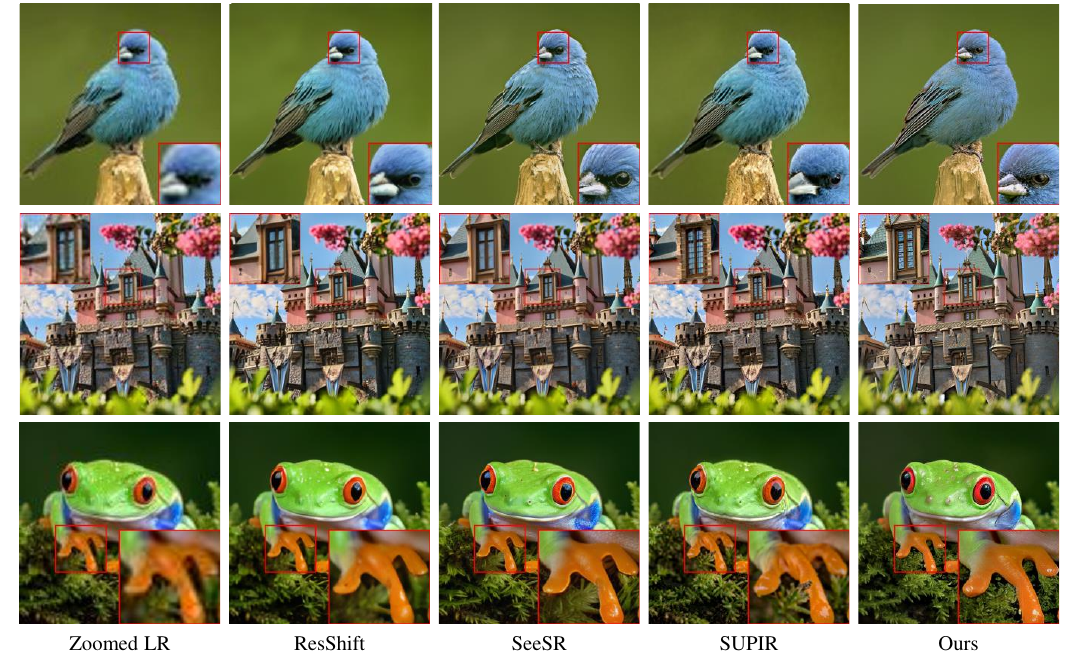}
	\caption{Visual comparisons between the proposed model and other state-of-the-art methods on Real-world datasets. For a clearer and more detailed view, please zoom in on the images.}
	\label{fig:visualcomp-real}
\end{figure*}

\subsubsection{Qualitative Comparisons}
We further provide visual comparisons of the proposed method against state-of-the-art Real-ISR methods on both synthetic and real-world images, as depicted in Figure~\ref{fig:visualcomp-DIV2k} and Figure~\ref{fig:visualcomp-real}.

\begin{figure*}[!t]
    \centering
    \includegraphics{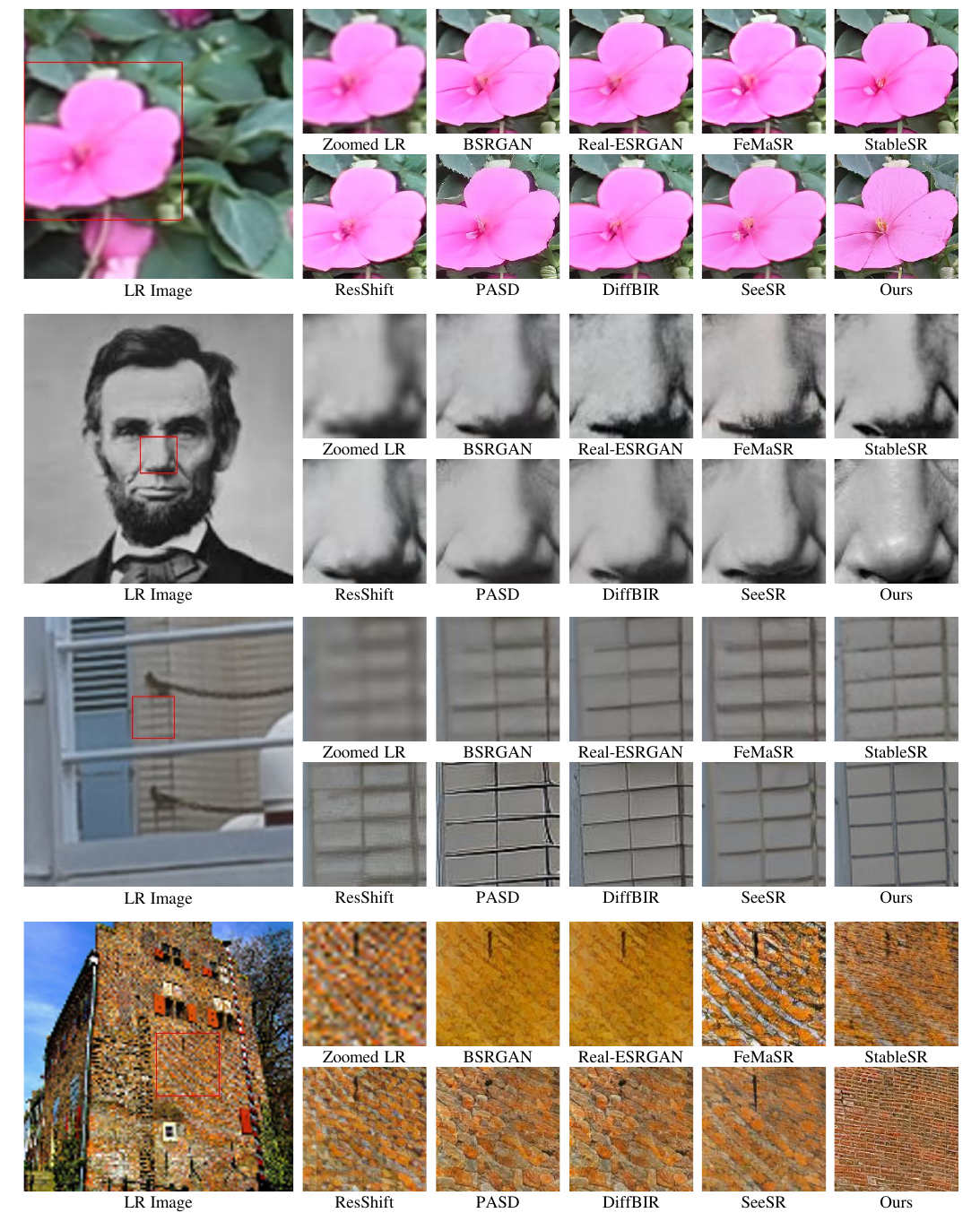}
    \caption{Qualitative comparisons with state-of-the-art methods on real-world examples are presented. For a closer examination of the finer details and to better appreciate the improvements, please zoom in on the images.}
    \label{fig:more-realresult}
\end{figure*}

\begin{figure*}[h]
	\centering
	\includegraphics[width= 0.95\textwidth]{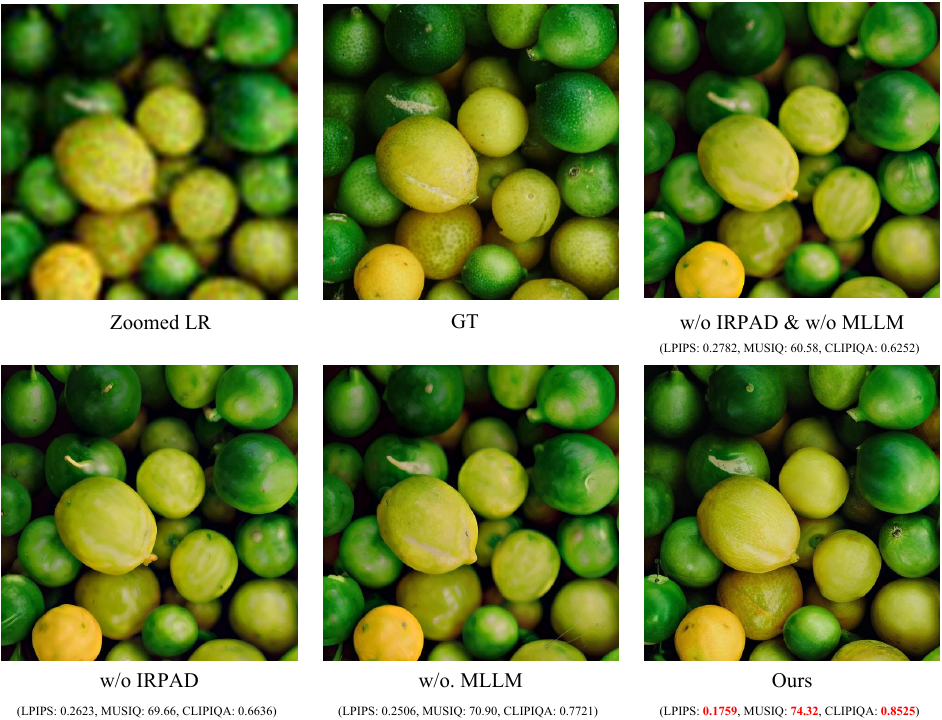}
	\caption{Visual comparison for ablation study. "w/o IRPAD \& w/o MLLM" represents model directly applying ControlNet on SD without degradation prompt and MLLM-generated semantic prompt. "w/o IRPAD" represents model utilizing MLLMs to generate semantic prompt, while without degradation prompt. "w/o MLLM" employs IRPAD to generate restoration prompts, while without MLLM-generated semantic prompt. "Ours" is final full model with both MLLM and IRPAD, resulting much clearer and more semantically accurate outcome. }
	\label{fig:ablation}
\end{figure*}

It is evident that all compared methods are capable of restoring visually high-quality images. Notably, the results produced by SeeSR and SUPIR exhibit finer details compared to those by ResShift, highlighting the significance of textual guidance in diffusion-based image restoration models. In Figure~\ref{fig:visualcomp-DIV2k}, our model tends to generate clearer and more regular contours, such as sharper leaves, finer animal fur, and well-defined flower petals. Additionally, in Figure~\ref{fig:visualcomp-real}, our model surpasses SeeSR and SUPIR in generating semantically accurate and detail-rich results, which highlights the advantages of using degradation-alignment prompt conditioning.

In summary, the proposed model demonstrates advantages in detail generation and degradation removal. These examples underscore the benefits of utilizing multimodal semantic priors over just relying on scattered object labels. More visual examples that further illustrate the capabilities of our model can be found in Figure~\ref{fig:more-realresult}.

\subsubsection{User Study}
\begin{figure}[h]
	\centering
	  \includegraphics[width=0.5\textwidth]
   {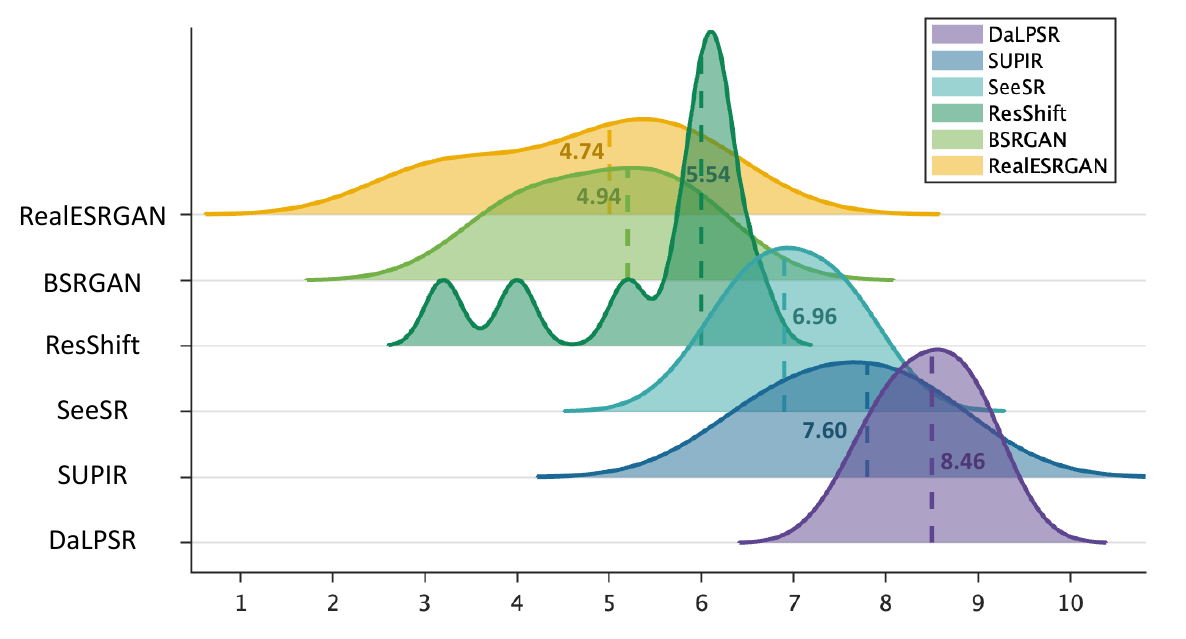}
        \caption{Comparisons about score distributions of user-study on real-world data for six representative state-of-the-art methods. The scores are divided into five levels. 2 for very poor, 4 for poor, 6 for fair, 8 for good, and 10 for excellent. }
	\label{fig:user-study}
\end{figure}

To substantiate the superiority of the proposed DaLPSR in generating precise and realistic LR images, we conducted a user study. This investigation involved 10 real-world LR images randomly selected from established benchmark datasets. We recruited 10 participants to evaluate six representative methods (BSRGAN, Real-ESRGAN, ResShift, SeeSR, SUPIR, and DaLPSR). As a result, each method can be valuated by 100 scores. During evaluation, the names of all method was concealed. Participants are blindly to rate "Which HR image best matches the LR reference and provides the best SR result?". Each HR image was rated on a scale: 2 for very poor, 4 for poor, 6 for fair, 8 for good, and 10 for excellent. As depicted in Figure~\ref{fig:user-study} and Table~\ref{tab:user-study}, our DaLPSR demonstrates a higher score compared to other methods when evaluated on real-world data. The user-study has further confirmed the advantages of our proposed DaLPSR.

\begin{table}[h]
\caption{Statistical comparisons of user study on real-world data for six representative state-of-the-art methods. }
\label{tab:user-study}
\tabcolsep 21pt
\begin{tabular*}{\linewidth}{c|c|c}
\toprule
Methods & Average Score & Best rates \\\hline
BSRGAN & 4.94 &  1\% \\
RealESRGAN & 4.74 & 3\% \\
ResShift & 5.54 & 4\% \\
SeeSR & 6.96 & 17\% \\
SUPIR & 7.60 & 30\% \\
DaLPSR & 8.46 & 45\% \\
\bottomrule
\end{tabular*}
\end{table}

\subsection{Ablation Studies}
To validate the effectiveness of our proposed key components, we have performed ablation studies on representative RealSR and RealLR200 datasets for practical use.
\subsubsection{Efficacy of Image-Restoration Prompt Alignment Decoder}
As shown in Table~\ref{tab:ablation}, the Image-Restoration Prompt Alignment Decoder (IRPAD) greatly improves the perceptual quality of restored low-resolution (LR) images, with enhancements evident across various evaluation metrics. This underscores the decoder's effectiveness in producing visually superior results. To provide a clearer understanding, we include visual comparisons from our ablation experiments. As depicted in Figure~\ref{fig:ablation}, the use of IRPAD significantly reduces image noise and minimizes aliasing artifacts, resulting in textures with well-defined edges. These improvements clearly demonstrate the decoder's capability to achieve visually appealing and high-fidelity restorations.

\begin{table}
\caption{Ablation studies on RealSR and RealLR200 benchmarks.}
\label{tab:ablation}
\tabcolsep 5.6pt
\begin{tabular*}{\linewidth}{cc|cc|cc}
\toprule
\multicolumn{2}{c}{Datasets} & \multicolumn{2}{c}{RealSR} & \multicolumn{2}{c}{RealLR200} \\\midrule
IRPAD & MLLM & MUSIQ$\uparrow$ & CLIPIQA$\uparrow$ & MUSIQ$\uparrow$ & CLIPIQA$\uparrow$ \\\hline
\text{\texttimes} & \text{\texttimes} & 66.72 & 0.6109 & 64.35 & 0.5916 \\
\text{\texttimes} & \checkmark & 68.54 & 0.6355 & 66.82 & 0.6495  \\
\checkmark & \text{\texttimes} & 69.69 & 0.6671 & 69.25 & 0.6750  \\
\checkmark & \checkmark & 71.48 & 0.7199 & 72.78 & 0.7728 \\
\bottomrule
\end{tabular*}
\end{table}

\subsubsection{Efficacy of Multi-Modality Semantic Priors}
As shown in Table~\ref{tab:ablation}, excluding high-level semantic priors results in a noticeable drop in fidelity metrics. This decrease emphasizes the essential role these priors play in maintaining the perceptual quality and structural integrity of restored images. Furthermore, as depicted in Figure~\ref{fig:ablation}, the effectiveness of using multi-modality semantic priors is clearly validated. Incorporating these priors leads to images with sharper details and more refined textures, enhancing realism and overall image quality.

In the ablation study, to ensure a fair comparison, we constructed baseline models by systematically excluding IRPAD and/or MLLM components. For the baseline without IRPAD, we removed the IRPAD module and relied on a basic image restoration pipeline involving standard denoising, lacking the refined adjustments provided by IRPAD. For the baseline without MLLM, we excluded MLLM components, processing images based solely on intrinsic features without multimodal inputs or semantic prompts, using only basic image characteristics for restoration. These baselines were trained and tested under identical conditions, using the same datasets and evaluation metrics such as LPIPS and CLIPIQA, to ensure comparability. By analyzing these models, we clearly demonstrate the specific contributions of IRPAD and MLLM to the overall restoration quality.

\begin{table}
\caption{Classification accuracy for different degradation types.}
\label{tab:degradationCLF}
\tabcolsep 22pt
\begin{tabular*}{\linewidth}{cccc}
\toprule
Degradation & Blur & Noise & JPEG  \\\midrule
Accuracy & 0.63 & 0.65 & 0.69 \\
\bottomrule
\end{tabular*}
\end{table}

\subsubsection{Discussion and Future Work}
In this work, we have innovatively simplified the generation of degradation prompt, and converted generation problem into isolate classification/retrieval problem for respective degradation type. As illustrated in ablation study, it plays significant role on model's performance improvement. Nevertheless, as detailed in Table~\ref{tab:degradationCLF}, the accuracy of four-class classification is less than 70\%, still unsatisfactory. We believe more accurate degradation prompt will be more benefit for restoration. In the future, we will further optimize the way of degradation prompt generation, and provide accurate controls to balance the generation quality and fidelity of HR image.

\section{Conclusion}
In this paper, we have proposed an effective and innovative multimodal framework that leverages degradation aligned language prompts for real-world image super-resolution. In the proposed framework, two complementary priors including semantic content priors and image degradation priors are produced as textual prompts to guide stable diffusion for high-fidelity high-resolution image generation. Comprehensive experiments on several popular synthetic and real-world benchmark datasets have demonstrated that the proposed method achieves significant superior performance, and obtains new state-of-the-art perceptual quality level, especially on real-world cases based on reference-free metrics. The proposed method offers a viable solution to overcome existing challenges with acceptable computation burdens.

\section*{Acknowledgment}
This work is supported by National Natural Science Foundation of China under Grand No. 62366021, and Jiangxi Provincial Graduate Innovation Funding Project under Grand YC2023-S298.

\bibliographystyle{IEEEtran}
\bibliography{references}

\end{document}